\documentclass[letterpaper, 10 pt, conference]{ieeeconf}  

\IEEEoverridecommandlockouts                              

\usepackage{graphicx}
\usepackage{amssymb}
\usepackage{amsmath}
\usepackage{array}
\usepackage{booktabs}
\usepackage{multirow}
\usepackage{subcaption}
\usepackage{algorithm2e}
\usepackage{tikz}
\usepackage{hyperref}

\title{\LARGE \bf
Pedestrian Environment Model for Automated Driving
}

\author{Adrian Holzbock$^{1}$, Alexander Tsaregorodtsev$^{1}$, and Vasileios Belagiannis$^{2}$
\thanks{$^{1}$Adrian Holzbock and Alexander Tsaregorodtsev are with the Institute of Measurement, Control and Microtechnology, Ulm University, 89081 Ulm, Germany 
        {\tt\small <firstname>.<lastname>@uni-ulm.de}
        }%
\thanks{$^{2}$Vasileios Belagiannis is with the Chair of Multimedia Communications and Signal Processing, Friedrich-Alexander-Universität Erlangen-Nürnberg, 91058 Erlangen, Germany
        {\tt\small vasileios.belagiannis@fau.de}}%
\thanks{Parts of this research have been conducted as part of the EVENTS project, which is funded by the European Union, under grant agreement No 101069614. Views and opinions expressed are however those of the author(s) only and do not necessarily reflect those of the European Union or European Commission. Neither the European Union nor the granting authority can be held responsible for them.}%
}

\begin{document}

\newcommand\copyrighttextinitial{
    \scriptsize This work has been submitted to the IEEE for possible publication. Copyright may be transferred without notice, after which this version may no longer be accessible.}
    
\newcommand\copyrighttextfinal{
    \scriptsize\copyright\ 2023 IEEE. Personal use of this material is permitted. Permission from IEEE must be obtained for all other uses, in any current or future media, including reprinting/republishing this material for advertising or promotional purposes, creating new collective works, for resale or redistribution to servers or lists, or reuse of any copyrighted component of this work in other works.}
    
\newcommand\copyrightnotice{
    \begin{tikzpicture}[remember picture,overlay]
    \node[anchor=south,yshift=10pt] at (current page.south) {{\parbox{\dimexpr\textwidth-\fboxsep-\fboxrule\relax}{\copyrighttextfinal}}};
    \end{tikzpicture}
}

\bstctlcite{IEEEexample:BSTcontrol}

\newcolumntype{C}[1]{>{\centering\arraybackslash}m{#1}}
\newcolumntype{L}[1]{>{\raggedright\arraybackslash}p{#1}}

\maketitle
\thispagestyle{empty}
\pagestyle{empty}

\begin{abstract}

Besides interacting correctly with other vehicles, automated vehicles should also be able to react in a safe manner to vulnerable road users like pedestrians or cyclists. For a safe interaction between pedestrians and automated vehicles, the vehicle must be able to interpret the pedestrian's behavior. Common environment models do not contain information like body poses used to understand the pedestrian's intent. In this work, we propose an environment model that includes the position of the pedestrians as well as their pose information. We only use images from a monocular camera and the vehicle's localization data as input to our pedestrian environment model. We extract the skeletal information with a neural network human pose estimator from the image. Furthermore, we track the skeletons with a simple tracking algorithm based on the Hungarian algorithm and an ego-motion compensation. To obtain the 3D information of the position, we aggregate the data from consecutive frames in conjunction with the vehicle position. We demonstrate our pedestrian environment model on data generated with the CARLA simulator and the nuScenes dataset. Overall, we reach a relative position error of around 16\% on both datasets.

\end{abstract}

\section{Introduction}
\label{sec:introduction}
\begin{figure}[ht]
    \centering
    \includegraphics[width=0.48\textwidth]{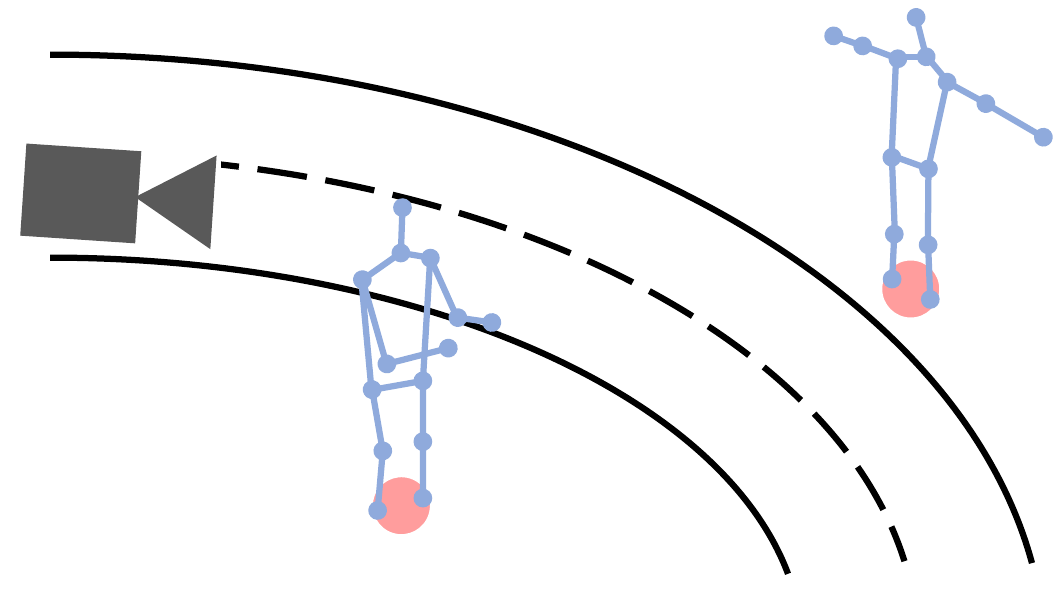}
    \caption{Pedestrian environment model visualization. Pedestrians are not only described by their position (displayed as a red dot) but also by their pose (shown as blue skeletons). The camera symbol illustrates the ego vehicle.}
    \label{fig:teaser}
\end{figure}

\copyrightnotice
In recent years, perception~\cite{9000872} and planning algorithms~\cite{9922601} made major contributions to automated driving. To safely navigate, the vehicle has to fully recognize the surrounding environment. This can be done with different environment models like grid maps~\cite{schreiber2022multi} and target lists~\cite{luo2021multiple}. Existing environment models are designed to display the location of other traffic participants like cars or pedestrians on the map and also show the driveable area but neglect the differing characteristics of pedestrians. Unlike cars, pedestrians usually communicate with other traffic participants by gestures, e.g., waving through a vehicle at a crosswalk or a police officer regulating the traffic~\cite{tcg_dataset}. In current environment models, important information for adequate and safe communication between the automated vehicle and the pedestrian, like poses, is missing. Therefore, we propose an environment model that contains both the pedestrian's position as well as its pose, as shown in Fig.~\ref{fig:teaser}. This enables the execution of gesture recognition~\cite{holzbock2022spatio}, human behavior understanding~\cite{9660784}, or body pose forecasting~\cite{ijcai2022p111} on our pedestrian environment model to better address the characteristics of pedestrians.

Common approaches to model the vehicle's environment are grid maps or track lists from multi-object tracking (MOT)~\cite{bar2001estimation}. Grid maps can be used to describe not only static environments~\cite{30720} but also surroundings with dynamic objects~\cite{lee2020pillarflow} and can even include the object's type~\cite{schreiber2022multi}. In MOT, the objects detected in the current frame are associated with objects from earlier frames to get their position and motion data. Bewley et al.~\cite{bewley2016simple} propose a simple algorithm for MOT in the 2D image plane. Other methods enable MOT in 3D space with lidar data~\cite{liang2022neural} or even improve the performance by fusing lidar and camera data~\cite{bai2022transfusion}. Both grid maps and track lists do not provide information about the pedestrian's pose. Our proposed pedestrian environment model includes, besides the pedestrian's position, also its pose in form of the human skeleton.

Our approach only needs data from a monocular camera sensor as well as a self-localization system to generate the pedestrian environment model. The method consists of three steps, namely body skeleton extraction, person tracking, and position estimation. During skeleton extraction, we extract 2D skeletons in pixel coordinates from each person visible in the image. For pose extraction, we use a pre-trained deep neural network for 2D body pose estimation. In the second step, we assign the extracted skeletons from the current frame to skeletons from previous frames and accumulate the pedestrians' pose information over time. Before the tracking step, we apply an ego-motion compensation algorithm to obtain more precise assignments during the tracking. To determine the 3D position of the person in a global coordinate system, we use the skeletal information of two consecutive frames and the self-localization data. We evaluate our pedestrian environment model on a dataset generated by CARLA~\cite{carla} and on the mini-set-split of the nuScenes dataset~\cite{nuscenes2019}. Our approach reaches promising performance in the position estimation of pedestrians for both datasets. To the best of our knowledge, we are the first to provide an environment model for pedestrians to improve the interaction modeling with automated vehicles. 

\section{Related Work}
\label{sec:related_work}
There are two common approaches to describe the automated vehicle's environment: Grid maps and target lists generated from multi-object tracking. 

\subsection{Grid Maps}
\label{subsec:grid_maps}
In classical occupancy grid maps~\cite{30720}, the vehicle's surrounding area is divided into single equally sized cells that can be free or occupied by another object. The cell states of the grid map can be updated with new measurements resulting in a measurement grid map. Therefore, an inverse sensor model processes the measurements and assigns an occupancy probability to the cell~\cite{thrun2002probabilistic}. Occupancy grid maps describe static environments but do not consider dynamic objects, which are common in automated driving. Dynamic occupancy grid maps include, besides the occupancy, also a velocity for the dynamic object. Tanzmeister et al.~\cite{tanzmeister2014grid} model dynamic elements in the grid map with particles. Schreiber et al.~\cite{9561375} apply recurrent neural networks to predict a dynamic occupancy grid map from measurement grids with an ego-motion compensation. An extension~\cite{schreiber2022multi} uses raw lidar point clouds as input and predicts additional semantic labels for the cells. While grid maps can provide information about free space for driving and other traffic participant attributes, like orientation, velocity, and type, they never included pedestrian body pose data. In contrast, the proposed pedestrian environment model can provide those details that are important for the communication between automated vehicles and pedestrians.

\subsection{Multi-Object Tracking}
\label{subsec:multi_object_tracking}
Multi-object tracking (MOT) delivers a track list of the different objects' locations in the vehicle's surroundings. We can divide the MOT into 2D tracking in the image plane and 3D tracking. A lightweight tracking algorithm in the image plane is the intersection over union tracker~\cite{fernandez2019real}, where the association is done with the intersection over union (IoU) between two frames. SORT~\cite{bewley2016simple} improves the simple IoU tracker by propagating the labels with the Kalman Filter~\cite{kalman_filter} to the next frame and resolves the assignment with the Hungarian Algorithm~\cite{hungarian_algorithm}. Instead of using the IoU as cost for the assignment, DeepSORT~\cite{wojke2017simple} takes the Mahalanobis distance and a feature vector produced from a pre-trained neural network. Using lidar data as a base for the MOT enables 3D MOT. Liang and Meyer~\cite{liang2022neural} introduce NEBP, which works with lidar data and complements belief propagation with graph neural networks. The fusion of multiple sensors like cameras and lidars is proposed to improve the MOT performance. EagerMOT~\cite{kim2021eagermot} fuses detections from 2D and 3D sensors and updates the tracks depending on the available detection information. TrackFormer~\cite{Meinhardt_2022_CVPR} combines the feature extraction and MOT in one neural network making separate object detectors unnecessary. The presented MOT methods track objects over time and provide a track list with other traffic participants for the planning of the automated vehicle. Our pedestrian environment model also tracks the pedestrians as the presented 2D approaches in the image plane but provides additional 3D world positions and skeletal information.

\section{Method}
\label{sec:method}
In the following section, we describe our method to generate a pedestrian environment model from the image of a monocular camera and the vehicle's self-localization system. 

\begin{figure*}[ht]
\vspace{2mm}
    \centering
    \includegraphics[width=0.96\textwidth]{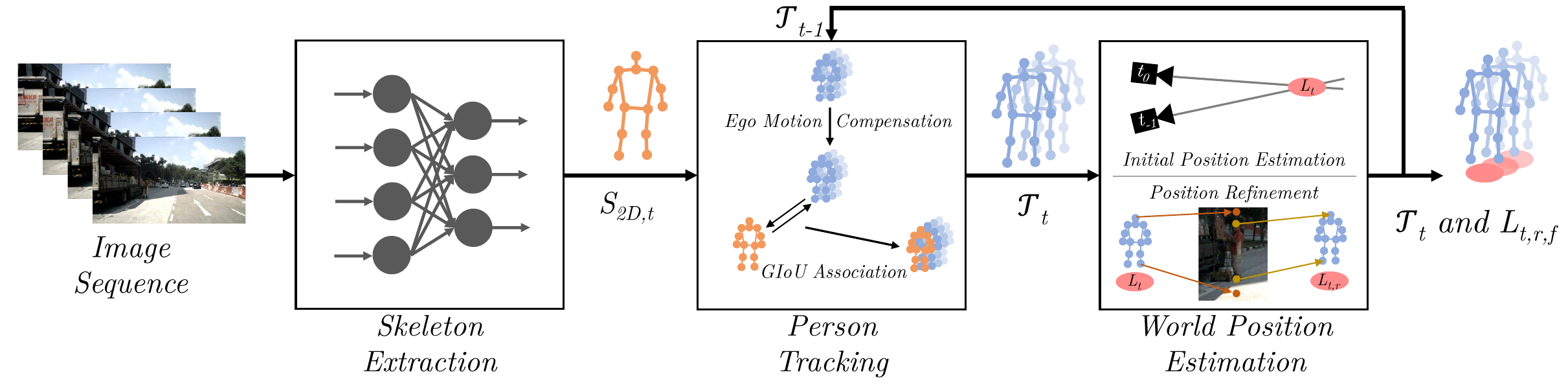}
    \caption{Overview of the proposed method to generate the pedestrian environment model. First, we extract from the image sequence the 2D skeletons $S_{2D,t}$ (orange skeleton) with a neural network. Then, we associate the skeletons from the current time step $S_{2D,t}$ with the track list from the previous time step $\mathcal{T}_{t-1}$ to get the updated track list $\mathcal{T}_{t}$ (sequence of blue skeletons). Finally, we estimate the pedestrians' 3D position in the world coordinate system $L_{t,r,f}$ (red dot).}
    \label{fig:overview}
\end{figure*}

\subsection{Method Overview}
\label{subsec:overview}
We aim to generate a pedestrian environment model by utilizing RGB images from a monocular camera sensor and the data of the vehicle's self-localization system. Finally, our environment model contains, besides the pedestrians' position in world coordinates, also their pose. The proposed method consists of three steps, visualized in Fig.~\ref{fig:overview}. In the first step (see Sec.~\ref{subsec:skeleton_extraction}), the pedestrian's skeletons and bounding boxes are extracted from the camera image using a neural network. The pedestrian's bounding boxes are then used in the second step (presented in Sec.~\ref{subsec:person_tracking}) to associate the detected pedestrians of the current time step with the detections from the last time step. To also obtain their position in the world coordinate frame, we calculate the position in the third step (described in Sec.~\ref{subsec:distance_identification}) based on the keypoints and the vehicle's self-localization information. Finally, we get the pose and the position in the world coordinate system of all pedestrians located in the camera's field of view.

\subsection{Skeleton Extraction}
\label{subsec:skeleton_extraction}
The pose information of the pedestrians relies only on the camera images. To extract the keypoints, we use CID~\cite{cid_pose}, a neural network pose estimator pre-trained on the COCO dataset~\cite{coco_dataset}. CID combines pedestrian detection and keypoint extraction in one neural network, which enables a constant execution time independent of the number of pedestrians. In contrast, the execution time using two separated neural networks, one for pedestrian detection and another for keypoint extraction, highly correlates with the number of pedestrians. Also, CID makes the keypoint grouping to a single pedestrian unnecessary because it predicts for each pedestrian all keypoints in an own heatmap. CID delivers a skeleton $S_{2D,t} = \{\mathbf{k}_{1,t}, \dots, \mathbf{k}_{i,t}, \dots, \mathbf{k}_{17,t}\}$ at time step $t$ that is described by the 2D pixel coordinates for the 17 keypoints $\mathbf{k}_{i,t} = [u, v]$ for all $m$ pedestrians in an image. Additionally, CID returns a detection probability vector $P_t = \{p_{1,t}, \dots, p_{i,t}, \dots, p_{17,t}\}$ with a detection probability $p_{i,t} \in \mathbb{R}^{1}$ for each keypoint. After keypoint extraction, the skeleton $S_{2D,t}$ is used to generate a 2D bounding box $b_t = \{S_{2D,t, u_\text{min}}, S_{2D,t, v_\text{min}}, S_{2D,t, u_\text{max}}, S_{2D,t, v_\text{max}}\}$, which is utilized in the pedestrian tracking in Sec.~\ref{subsec:person_tracking}. After this step, each pedestrian in the image is defined by a 2D skeleton $S_{2D, t}$, a detection probability $P_t$, and a bounding box $b_t$.

\subsection{Person Tracking}
\label{subsec:person_tracking}
To reconstruct temporal dependencies, the skeletons of the current time step must be associated with skeletons from earlier frames. In this step, we do not have any information about the distance to the pedestrian. Therefore, we use a simple tracking algorithm in the 2D image plane. In addition to the pedestrians' motion, the ego-motion of the vehicle can also lead to object shifts between the image frames. As we assume that pedestrian recognition is only necessary for urban traffic, we expect a speed limit of 50 km/h. Therefore, we neglect the shift between image frames due to the ego translation, as at this speed, an object's location shifted only by the perspective change still overlaps in two consecutive frames. We focus on the compensation of the ego vehicle's rotation (yaw), which can lead to huge perspective shifts between image frames. For the yaw ($\Psi$) compensation, we use the change of $\Psi$ between two time steps $\Delta \Psi$ from the ego localization system, the aperture angle of the camera $\alpha$, and the width of the image $w$. The yaw compensation of the 2D skeleton $S_{2D,t}$ and the bounding boxes $b_t$ only affects the $u$-value of the pixel coordinate $[u,v]$ and can be expressed as follows:
\begin{equation}
\label{eq:yaw_compensation}
    u_\text{comp} = u - \frac{\Delta \Psi}{\alpha} * w.
\end{equation}

During pedestrian tracking, we hold $n$ pedestrian tracks $\mathcal{T}_{t-1} = \{T_{0,t-1}, \dots, T_{n,t-1}\}$, where each track $T$ contains bounding boxes $b$ as well as 2D skeletons $S_{2D}$ from the last time steps. Before the association between the new pedestrians and the tracks, we compensate the track data with Eq.~\ref{eq:yaw_compensation}. Then, we calculate the generalized intersection over union (GIoU)~\cite{giou} between the pedestrians' bounding boxes $b_t$ and the tracks' last bounding boxes $b_{t-1}$. If a pedestrian from the current time step has no intersection with an existing track, we set up a new track for this pedestrian and do not consider them in the association. Besides the new pedestrians, we exclude tracks from the association that do not overlap with new pedestrians and remove them after three failed association attempts. By excluding new pedestrians and tracks with missing detections during the association, the number of new pedestrians $m$ and tracks $n$ is equal. We apply the Hungarian algorithm~\cite{hungarian_algorithm} to assign one person to each track such that the association cost $\mathbf{c} \in \mathbb{R}^{m \times n}$ is minimized as follows
\begin{equation}
    \mathrm{min}\sum_{j=0}^{m-1} \sum_{k=0}^{n-1} \mathbf{c}_{j,k} \mathbf{a}_{j,k},
\end{equation}

where the cost $\mathbf{c}_{j,k}$ is the negative GIoU and $\mathbf{a}_{j,k} = 1$, if the pedestrian $j$ is associated to track $k$. The outcome of the tracking is an updated list of tracks $\mathcal{T}_{t}$ improved with the pedestrians' states of the current time step containing the ego-motion compensated skeletons $S_{2D,t}$ and bounding boxes $b_t$ in pixel coordinates.

\subsection{World Position Estimation}
\label{subsec:distance_identification}
Besides the pose, the pedestrian's position related to the ego vehicle includes essential information. The image from the monocular camera provides a high angular resolution which can be used to determine the pedestrian's direction. In contrast, the details provided by the camera are insufficient to calculate the distance between the pedestrian and the vehicle. We propose a two-step method to identify the pedestrian's world position using the vehicle's self-localization and two consecutive image frames. First, we estimate an initial position by using geometric dependencies between the frames. Following, we refine the position using the pedestrian's image height. An overview of our world position estimation is given in Algorithm~\ref{alg:distance_estimation}.

\paragraph{Initial Estimation}
In the first step, we determine the initial world location of the pedestrian by using geometric dependencies. Therefore, we convert the 2D skeleton $S_{2D,t}$ from pixel coordinates into normalized 3D vectors in the world coordinate system. The 3D vector for a keypoint defines a ray in the 3D space on which the keypoint is located, while the distance is unknown. The conversion of a single 2D keypoint $\mathbf{k}_{i,t} = [u, v, 1]^T$ of time step $t$ to a normalized 3D direction vector in world coordinates $\mathbf{d}_{i,t} \in \mathbb{R}^3$ can be expressed as follows
\begin{equation}
\label{eq:direction_vector}
    \mathbf{d}_{i,t} = \mathbf{R}_\text{ext} \frac{\mathbf{K}_\text{int} \mathbf{k}_{i,t}}{||\mathbf{K}_\text{int} \mathbf{k}_{i,t}||}.
\end{equation}
Here, $\mathbf{K}_\text{int} \in \mathbb{R}^{3 \times 3}$ is the intrinsic camera calibration matrix and $\mathbf{R}_\text{ext} \in \mathbb{R}^{3 \times 3}$ the extrinsic camera rotation matrix. By converting the 2D skeleton $S_{2D,t}$, we obtain normalized 3D direction vectors $D_{3D,t} = \{\mathbf{d}_{1,t}, \dots, \mathbf{d}_{i,t}, \dots, \mathbf{d}_{17,t}\}$ for each keypoint. To determine the distance of a specific keypoint, we calculate the closest point between the rays of two consecutive frames by using the 3D direction vector $\mathbf{d}_{i,t}$ and the vehicle's ego position $\mathbf{o}_t$. From both, we use the data of the current time step $t$ and the previous time step $t-1$. The closest point on the ray of the current time step $\mathbf{o}_t + l_{i,t} \mathbf{d}_{i, t}$ and the previous time step $\mathbf{o}_{t-1} + l_{i,t-1} \mathbf{d}_{i, t-1}$ is defined by the following equation:
\begin{equation}
\label{eq:intersection}
    \mathbf{o}_t + l_{i,t} \mathbf{d}_{i, t} + l_{s,i} (\mathbf{d}_{i, t-1} \times \mathbf{d}_{i, t}) = \mathbf{o}_{t-1} + l_{i,t-1} \mathbf{d}_{i, t-1}.
\end{equation}
$l_{i,t}$ and $l_{i,t-1}$ define the distance from the origin $\mathbf{o}_t$ and $\mathbf{o}_{t-1}$ to the closest point on the corresponding ray, while $l_{s,i}$ represents the shortest distance between the rays. As the final estimate for the keypoint $\mathbf{k}_{3D, i,t}$, we use the midpoint between the two closest points defined by
\begin{equation}
\label{eq:mean_keypoints}
    \mathbf{k}_{3D, i,t} = \frac{\mathbf{o}_t + l_{i,t} \mathbf{d}_{i, t} + \mathbf{o}_{t-1} + l_{i,t-1} \mathbf{d}_{i, t-1}}{2},
\end{equation}
where $l_{i,t}$ and $l_{i,t-1}$ are obtained by solving Eq.~\ref{eq:intersection}. We combine the 3D keypoints $\mathbf{k}_{3D, i,t}$ to a 3D skeleton $S_{3D,t} = \{\mathbf{k}_{3D, 1,t}, \dots, \mathbf{k}_{3D, i,t}, \dots, \mathbf{k}_{3D, 17,t}\}$. To get the pedestrian's position in world coordinates, we calculate the mean of all 3D keypoints $\mathbf{k}_{3D, i,t}$. Assuming that the distance to keypoints $\mathbf{k}_{3D, i,t}$ with a low detection probability $p_{i,t}$ is inaccurate, we exclude 30\% of the keypoints with the lowest detection probability in the calculation of the pedestrian's world position. Finally, with the geometric dependencies, we get an initial estimation of the person's location $L_t = [x, y, z]^T$ in the world coordinate system.

\paragraph{Position Refinement}
Different factors like pedestrians' motion, inaccuracies in the pose estimation network, or a standing ego vehicle can reduce the accuracy of the pedestrian's initial position estimation $L_t$. In the second step, we refine the distance and direction of the pedestrian's position. For the distance refinement, we re-project the bottom and top point of the pedestrian to the image while we set the bottom point to 0.0 m and the top point to ${\text{1.7 m}}$, assuming an average person's height of 1.7 m. The re-projection of a point in world coordinates to the image plane can be conducted with the inverse intrinsic camera matrix $\mathbf{K}_{int}^{-1}$ and the inverse extrinsic camera matrix consisting of the rotation matrix $\mathbf{R}_{ext}$ and the translation vector $\mathbf{t} \in \mathbb{R}^3$ with the following equation:
\begin{equation}
\label{eq:reproject}
    \begin{bmatrix}
        u_r\\v_r\\1
    \end{bmatrix}
    = \mathbf{K}_\text{int}^{-1} 
    \begin{bmatrix}
        \mathbf{R}_\text{ext} & \mathbf{t}\\
        0 & 1 \\
    \end{bmatrix}^{-1}
    \begin{bmatrix}
        x\\y\\z\\1
    \end{bmatrix}.
\end{equation}
We use the re-projected top and bottom point to determine the person's height $h_\text{est}$ in pixels with the initial estimated distance. If the initially estimated distance corresponds to the ground truth distance, the difference between the re-projected $h_\text{est}$ and the original image height $h_\text{orig}$ in pixels is zero. During the distance refinement, we use the original and re-projected height to calculate a scaling factor that re-scales the distance between the pedestrian and the ego vehicle. The refined distance $l_r$ is calculated as follows:
\begin{equation}
\label{eq:refine}
    l_{r} = l * \biggl( 1+ \Bigl(\frac{h_\text{est}}{h_\text{orig}} - 1\Bigr) * \lambda \biggr).
\end{equation}
Here, $l$ defines the initial distance estimation and $\lambda$ is a dynamic scaling factor. We use the refined distance $l_r$ to update the person's position. In addition to the person's distance from the camera sensor, we refine their direction. For the direction refinement, we calculate the mean direction of the 3D direction vector $D_{3D,t}$ and set this as the updated direction.

To increase the accuracy, we repeat the refinement of the distance and the direction 15 times and progressively reduce the scaling factor $\lambda = r + 5$ for $r \in [1, \dots, 15]$. From the position refinement, we get the improved pedestrian's position $L_{t,r}$, which we smooth with a Kalman filter~\cite{kalman_filter} based on a constant velocity model. The filtering eliminates the influence of outliers and improves the position $L_{t,r,f}$.

\RestyleAlgo{ruled}
\SetKwInput{KwInput}{Input}
\SetKwInput{KwOutput}{Output}
\SetKwComment{Comment}{/* }{ */}
\begin{algorithm}[!ht]
\caption{Overview of the position estimation}
\label{alg:distance_estimation}
\DontPrintSemicolon
  
\KwInput{Current 2D keypoints $\mathbf{k}_{i,t}$, previous 2D keypoints $\mathbf{k}_{i,t-1}$, current ego position $\mathbf{o}_t$, previous ego position $\mathbf{o}_{t-1}$, intrinsic camera matrix $\mathbf{K}_{int}$, extrinsic rotation matrix $\mathbf{R}_{ext}$, extrinsic translation vector $\mathbf{t}$, number of refinement steps $s = 15$, scaling factor constant $c = 5$}
\KwOutput{3D pedestrian position $L_{t,r,f}$}
    \Comment*[l]{Initial distance estimation}
    Calculate direction vector $\mathbf{d}_{i,t}$ for $\mathbf{k}_{i,t}$ with Eq.~\ref{eq:direction_vector}\\
    Calculate 3D keypoint $\mathbf{k}_{3D,i,t}$ with Eq.~\ref{eq:intersection} and Eq.~\ref{eq:mean_keypoints}\\
    Calculate persons location $L_t$ as the mean of $\mathbf{k}_{3D,i,t}$
    \Comment*[l]{Distance Refinement}
    \For{$r=0$ to s} 
    {
    Re-project persons height to image using Eq.~\ref{eq:reproject}\\
    Refine distance with Eq.~\ref{eq:refine} and $\lambda = r + c$\\
    Refine angle with the mean direction of $\mathbf{k}_{3D,i,t}$
    }
    Filter $L_{t,r}$ with the Kalman Filter to get the final position $L_{t,r,f}$

\end{algorithm}

\section{Evaluation}
\label{sec:evaluation}
We evaluate our method on simulated and real-world traffic scenarios. In the following, we explain the utilized datasets, our experimental settings, and present the results.

\subsection{Datasets}
\label{subsec:datasets}
The experiments are conducted on a dataset generated with the CARLA simulator~\cite{carla} and the real-world dataset nuScenes~\cite{nuscenes2019}. Both datasets contain diverse scenes, including scenarios with crossing pedestrians in front of the car and pedestrians on the sidewalk.

\paragraph{CARLA Dataset}
We simulate different traffic scenes with the CARLA simulator~\cite{carla}. CARLA is an open-source simulator for automated driving that supports different environments and commonly used sensors in automated driving. Furthermore, CARLA can generate matching labels for the sensor data, e.g., bounding boxes, semantic segmentation maps, depth maps, and even pedestrians' skeletons. The scenes in the simulator can be generated automatically with vehicles driven by an autopilot and pedestrians controlled by an artificial intelligence. We generate the sequences with a simulation frequency of 10 Hz and place an RGB monocular camera next to the rear-view mirror. As camera parameters, we set the image resolution to $1600 \times 900$ pixels and the opening angle to $64.5$ degrees. In addition to the camera image, we simulate the self-localization data and labels, which contain 2D skeletons, 2D bounding boxes, and 3D bounding boxes. We filter out labels that are either outside the camera's field of view or occluded by other objects. To detect occlusions, we simulate a depth map and check if the pedestrian is visible.

\paragraph{nuScenes Dataset}
nuScenes~\cite{nuscenes2019} is a dataset for automated driving collected in Boston and Singapore under different environmental conditions. The dataset contains data from the entire sensor suite of an automated vehicle, including camera, lidar, radar, and localization data. Overall, the dataset consists of 1000 manually selected scenes with a length of 20 seconds labeled with 3D bounding boxes and object class labels at a frequency of 2 Hz. To evaluate our proposed method, we take the nuScenes mini-set-split, which contains a subset of 10 sequences. From this subset, we use the day scenes (Scenes: 61, 103, 533, 916, 655, 757, 796) for the evaluation and exclude the remaining night scenes as the used human pose estimator is not trained on night view images. In the evaluation, we take the images from the vehicle's front camera, which have a resolution of $1600 \times 900$ pixels, an opening angle of $64.5$ degrees, and a frame rate of 12 Hz. Furthermore, we use the labels of all pedestrians in the camera's field of view and the self-localization system data.

\subsection{Experimental Settings}
\label{subsec:dataset}
Following, we give an overview of our implementation\footnote{Our code is publicly available at \url{https://github.com/holzbock/ped\_env}.}. During the evaluation on the nuScenes dataset, we use the CID~\cite{cid_pose} human pose estimation neural network to extract the pedestrians' skeletons from the images. The network has an HRNet-W32~\cite{hrnet} as backbone and the parameters are trained on the COCO keypoint dataset~\cite{coco_dataset}. Due to the domain gap between real images and the images simulated with CARLA, we use the generated 2D skeletons to evaluate on the CARLA dataset instead of extracting them with a neural network. Further, using generated 2D skeletons makes the evaluation independent of the pose estimator's performance. Our approach applies the 2D skeletons and the self-localization data to create the pedestrian environment model. In the evaluation, we report the absolute error in meters $e_\text{abs}$ between the ground truth position and the predicted position and the relative error $e_\text{rel}$ to the ground truth distance. To assign a ground truth label to a pedestrian in the environment model, we first calculate the intersection over union between the ground truth and predicted pedestrians in the image plane. Then, we use the ground truth label with an intersection in the image plane closest to the world coordinate system.

\subsection{Results}
\label{subsec:results}
We evaluate our method on simulated CARLA data and nuScenes data with the described experimental settings.

\begin{figure}
\centering
\begin{subfigure}{.485\textwidth}
  \centering
  \includegraphics[width=0.95\linewidth]{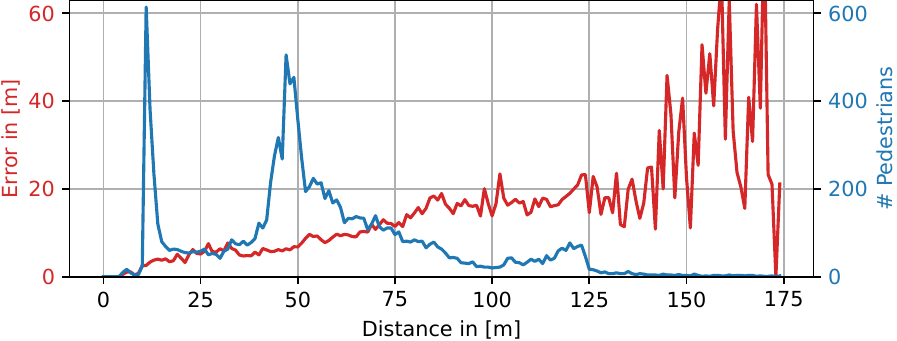}
  \caption{CARLA Simulated Dataset}
  \label{fig:carla_error_distance}
\end{subfigure}
\begin{subfigure}{.485\textwidth}
  \centering
  \includegraphics[width=0.95\linewidth]{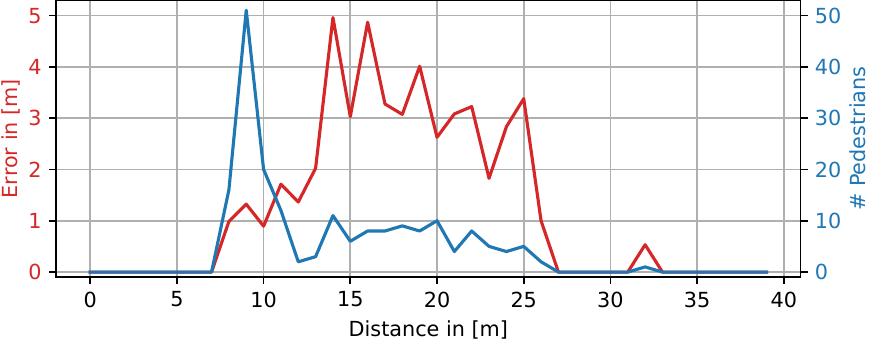}
  \caption{nuScenes Dataset}
  \label{fig:nuscene_error_distance}
\end{subfigure}
\caption{Absolute error plotted over the distance between vehicle and pedestrian for the CARLA dataset in the top image and for the nuScenes dataset in the bottom image.}
\label{fig:error_distance}
\end{figure}

\paragraph{CARLA Dataset}
In Table~\ref{tab:results_carla}, we present the results of our method on the simulated CARLA dataset, where the mean distance to pedestrians is around 57 m. The first row shows the overall error, which is in absolute numbers 9.123 m and relative to the ground truth distance 16.85\%. The scenes with the highest error are \textit{CARLA 1} and \textit{CARLA 3}, where the vehicle is waiting at a red traffic light and pedestrians are crossing in front of the vehicle. The best results are reached in the scene \textit{CARLA 5}. Here, the ego vehicle is in motion and the pedestrians are crossing the street at a greater distance to the sensor. Additionally, to the results in Table~\ref{tab:results_carla}, we provide the distance-depending error between the pedestrian and the ego vehicle in Fig.~\ref{fig:carla_error_distance}. As displayed in Fig.~\ref{fig:carla_error_distance}, the error increases with a higher distance between the pedestrian and the ego vehicle. Beginning at a distance of 125 m, the error increases even further and seems very noisy. A reason for the noise could be that fewer pedestrians are visible at high distances to the sensor, therefore outliers have a strong influence.

\begin{table}[]
\vspace{1mm}
  \caption{Evaluation results for our method on the CARLA dataset. \textit{CARLA all} is the mean result over all scenes, while the other rows show the mean result of one scene.}
  \label{tab:results_carla}
  \begin{center}
  {\small{
  \begin{tabular}{L{1.75cm}C{1.95cm}C{1.9cm}C{1.2cm}}
    \toprule
    Data & Absolute Error $e_\text{abs}$ in [m] & Relative Error $e_\text{rel}$ in [\%] & \# Pedestrians \\
    \midrule
    CARLA all & 9.123 & 16.85 & 12771 \\
    CARLA 0 & 11.970 & 18.54 & 963 \\
    CARLA 1 & 6.823 & 18.75 & 3161 \\
    CARLA 2 & 7.607 & 13.94 & 316 \\
    CARLA 3 & 11.922 & 18.74 & 3415 \\
    CARLA 4 & 7.649 & 12.82 & 459 \\
    CARLA 5 & 2.023 & 4.45 & 310 \\
    CARLA 6 & 9.112 & 16.84 & 201 \\
    CARLA 7 & 9.521 & 14.92 & 1642 \\
    CARLA 8 & 9.055 & 16.14 & 1942 \\
    CARLA 9 & 3.056 & 8.68 & 362 \\
    \bottomrule
\end{tabular}
}}
\end{center}
\end{table}

\paragraph{nuScenes Dataset}
The results for the evaluation on the nuScenes dataset are presented in Table~\ref{tab:results_nuscenes} and Fig.~\ref{fig:nuscene_error_distance}. The mean distance between the ego vehicle and the pedestrians in the nuScenes dataset is around 15 m. The overall result on the nuScenes dataset for all selected scenes is given in the first row of Table~\ref{tab:results_nuscenes} and is in absolute numbers 2.524 m and relative to the ground truth distance 15.66\%. In the scenes \textit{nuScenes 655}, \textit{nuScenes 757}, and \textit{nuScenes 796} almost no pedestrians are detected, which is why no further conclusions are made here. The best performance is reached on \textit{nuScenes 553}, where the vehicle waits at an intersection and pedestrians are crossing. The highest error occurs in \textit{nuScenes 103}, where the vehicle is driving through a street and pedestrians walk on the sidewalk. As in the evaluation of the synthetic data, we provide an overview of the error in dependency on the distance between the pedestrian and the ego vehicle in Fig.~\ref{fig:nuscene_error_distance}. Here, we can see that we reach an error below 2 m up to a distance of around 13 m. Afterward, the error rises, but the number of detected pedestrians in distances above 13 m decreases.

\begin{table}[]
\vspace{1mm}
  \caption{Evaluation results for our method on nuScenes. \textit{nuScenes all} is the mean result over all scenes, while the other rows show the mean result of one scene.}
  \label{tab:results_nuscenes}
  \begin{center}
  {\small{
  \begin{tabular}{L{1.85cm}C{1.95cm}C{1.9cm}C{1.2cm}}
    \toprule
    Data & Absolute Error $e_\text{abs}$ in [m] & Relative Error $e_\text{rel}$ in [\%] & \# Pedestrians \\
    \midrule
    nuScenes all & 2.524 & 15.66 & 194 \\
    nuScenes 61 & 3.032 & 18.75 & 29 \\
    nuScenes 103 & 4.232 & 20.10 & 25 \\
    nuScenes 553 & 1.809 & 13.21 & 96 \\
    nuScenes 655 & 3.129 & 16.01 & 4 \\
    nuScenes 757 & 0.000 & 0.00 & 0 \\
    nuScenes 796 & 9.270 & 62.21 & 1 \\
    nuScenes 916 & 2.579 & 15.33 & 39 \\
    \bottomrule
\end{tabular}
}}
\end{center}
\end{table}

\subsection{Ablation Study}
\label{subsec:ablation}
The following section presents further investigations to give a deeper insight into our approach. We evaluate the influence of the refinement step, the effect of using multiple points for the initial distance estimation, and measure the run-time of the whole approach. The experiments in this section are all performed on the nuScenes dataset to show the influence of real data.

\paragraph{Distance Refinement}
We use the refinement step to improve the position accuracy, which can be inaccurate due to noisy pose estimates or a standing ego vehicle. On the nuScenes data, we reach an overall absolute error of 2.524 m with the distance refinement step. By skipping the refinement step and only using the initial position estimate, the error increases to 6.601 m, representing a relative error of over 50\%. The effect of the missing refinement step can especially be seen in \textit{nuScenes 553}, where the ego vehicle is standing at a red traffic light. In this case, the refinement step can improve the relative error from 73.93\% to 13.21\%.

\paragraph{Single Point Initial Distance}
In the initial estimation of the pedestrian's position, we use an average of all keypoints to determine the position. In this ablation study, we only use the midpoint of the pedestrian to calculate the initial pedestrian's position. By using all keypoints for the initial estimation, we can improve the error from 2.562 m when using only the midpoint to 2.524 m when using all keypoints. The difference between the midpoint and all keypoints is even higher when the refinement step is skipped. Then, using all keypoints for the initial estimate gives an overall error of 6.601 m and with only the midpoint, the error increases to 7.890 m. This shows that using all keypoints for the initial distance estimate is beneficial, as outliers can be suppressed.

\paragraph{Method Run-Time}
In automated driving, computational overhead plays a significant role due to limited resources. Therefore, we also measure the run-time of our approach. For a more meaningful result, we measure the run-time on the nuScenes dataset 5 times and calculate the mean. We determine the run-time of our approach implemented in Python on a workstation with an AMD Threadripper 3960x, 64 GB RAM, and an Nvidia GeForce RTX 3090. The pose estimation network is executed with PyTorch~\cite{pytorch} on the GPU, while the remaining code runs on the CPU. The overall mean run-time of our approach is 40.54 ms, of which the neural network for the pose estimation causes 39.94 ms. Our simple tracking approach takes 0.35 ms and the position calculation needs 0.29 ms per pedestrian. The neural network and the tracking have a constant run-time independent of the number of detected pedestrians. In contrast, the run-time of the position calculation depends on the number of pedestrians and, on average, needs 0.29 ms for every detected pedestrian.

\subsection{Result Discussion}
\label{subsec:result_discussion}
Our experiments show that the proposed method can reach promising results in generating a pedestrian environment model. We reach a distance error below 2 m close to the ego vehicle by only using the data from a monocular camera and the self-localization system. Compared to the simulated dataset, the nuScenes dataset has fewer samples during evaluation. This is caused, on the one hand, due the fact that we can evaluate only every sixth frame because of the sparse labels and, on the other hand, by the performance of the human pose estimator. The human pose estimator only detects pedestrians up to a distance of around 25 m, which can also be seen in Fig.~\ref{fig:nuscene_error_distance}. The fact that the pose estimation network detects pedestrians only in a range of up to 25 m also causes the difference in the overall absolute error between the simulated dataset and the nuScenes dataset, while the relative error is equal for both datasets. Furthermore, on the simulated dataset, the lowest performance is reached on a scene where the vehicle is waiting at a red traffic light where pedestrians are crossing. In this situation, on the nuScenes dataset, the best performance is reached. Erroneous assignments cause this during the tracking of the pedestrians in the simulation in cases pedestrians occlude further away pedestrians. This does not happen in real scenarios because of the limited performance of the human pose estimator.

\section{Conclusions}
\label{sec:conclusions}
This work presented a pedestrian environment model generated only by RGB monocular camera images and self-localization data. In addition to the positional information, the proposed pedestrian environment model also contains the human pose, which enables downstream tasks like human gesture recognition and human behavior understanding. Our approach consists of three different steps. In the first step, we extract the human skeletons from the RGB image. In the next step, we compensate the ego-motion and associate the skeletons from the current time step to earlier time steps. Lastly, we determine the pedestrian's position in world coordinates. Therefore, we use the self-localization data and the skeletons from two consecutive time steps. Further, we refine the position by re-projecting the pedestrian's height to the image. We evaluate our pedestrian environment model on a synthetic dataset generated with the CARLA simulator, as well as the real-world dataset nuScenes. 


\bibliographystyle{ieeetran}
\bibliography{references}

\begin{thebibliography}{10}
\providecommand{\url}[1]{#1}
\csname url@rmstyle\endcsname
\providecommand{\newblock}{\relax}
\providecommand{\bibinfo}[2]{#2}
\providecommand\BIBentrySTDinterwordspacing{\spaceskip=0pt\relax}
\providecommand\BIBentryALTinterwordstretchfactor{4}
\providecommand\BIBentryALTinterwordspacing{\spaceskip=\fontdimen2\font plus
\BIBentryALTinterwordstretchfactor\fontdimen3\font minus
  \fontdimen4\font\relax}
\providecommand\BIBforeignlanguage[2]{{%
\expandafter\ifx\csname l@#1\endcsname\relax
\typeout{** WARNING: IEEEtran.bst: No hyphenation pattern has been}%
\typeout{** loaded for the language `#1'. Using the pattern for}%
\typeout{** the default language instead.}%
\else
\language=\csname l@#1\endcsname
\fi
#2}}
\renewcommand\BIBentryALTinterwordstretchfactor{4}

\bibitem{9000872}
D.~Feng, \emph{et~al.}, ``Deep multi-modal object detection and semantic
  segmentation for autonomous driving: Datasets, methods, and challenges,''
  \emph{IEEE Transactions on Intelligent Transportation Systems}, vol.~22,
  no.~3, pp. 1341--1360, 2021.

\bibitem{9922601}
L.~Peng, Y.~Yan, J.~Wang, D.~Han, Y.~Yao, and G.~Yin, ``Hierarchical motion
  planning system with consideration of the dynamic lane-changing behaviour,''
  in \emph{2022 IEEE 25th International Conference on Intelligent
  Transportation Systems (ITSC)}, 2022, pp. 3455--3460.

\bibitem{schreiber2022multi}
M.~Schreiber, V.~Belagiannis, C.~Gl{\"a}ser, and K.~Dietmayer, ``A multi-task
  recurrent neural network for end-to-end dynamic occupancy grid mapping,'' in
  \emph{2022 IEEE Intelligent Vehicles Symposium (IV)}.\hskip 1em plus 0.5em
  minus 0.4em\relax IEEE, 2022, pp. 315--322.

\bibitem{luo2021multiple}
W.~Luo, J.~Xing, A.~Milan, X.~Zhang, W.~Liu, and T.-K. Kim, ``Multiple object
  tracking: A literature review,'' \emph{Artificial intelligence}, vol. 293, p.
  103448, 2021.

\bibitem{tcg_dataset}
J.~Wiederer, A.~Bouazizi, U.~Kressel, and V.~Belagiannis, ``Traffic control
  gesture recognition for autonomous vehicles,'' in \emph{2020 IEEE/RSJ
  International Conference on Intelligent Robots and Systems (IROS)}.\hskip 1em
  plus 0.5em minus 0.4em\relax IEEE, 2020, pp. 10\,676--10\,683.

\bibitem{holzbock2022spatio}
A.~Holzbock, A.~Tsaregorodtsev, Y.~Dawoud, K.~Dietmayer, and V.~Belagiannis,
  ``A spatio-temporal multilayer perceptron for gesture recognition,'' in
  \emph{2022 IEEE Intelligent Vehicles Symposium (IV)}.\hskip 1em plus 0.5em
  minus 0.4em\relax IEEE, 2022, pp. 1099--1106.

\bibitem{9660784}
M.~Herman, \emph{et~al.}, ``Pedestrian behavior prediction for automated
  driving: Requirements, metrics, and relevant features,'' \emph{IEEE
  Transactions on Intelligent Transportation Systems}, vol.~23, no.~9, pp.
  14\,922--14\,937, 2022.

\bibitem{ijcai2022p111}
A.~Bouazizi, A.~Holzbock, U.~Kressel, K.~Dietmayer, and V.~Belagiannis,
  ``Motionmixer: Mlp-based 3d human body pose forecasting,'' in
  \emph{Proceedings of the Thirty-First International Joint Conference on
  Artificial Intelligence, {IJCAI-22}}, 7 2022, pp. 791--798.

\bibitem{bar2001estimation}
Y.~Bar-Shalom, X.~R. Li, and T.~Kirubarajan, \emph{Estimation with applications
  to tracking and navigation: theory algorithms and software}.\hskip 1em plus
  0.5em minus 0.4em\relax John Wiley \& Sons, 2001.

\bibitem{30720}
A.~Elfes, ``Using occupancy grids for mobile robot perception and navigation,''
  \emph{Computer}, vol.~22, no.~6, pp. 46--57, 1989.

\bibitem{lee2020pillarflow}
K.-H. Lee, \emph{et~al.}, ``Pillarflow: End-to-end birds-eye-view flow
  estimation for autonomous driving,'' in \emph{2020 IEEE/RSJ International
  Conference on Intelligent Robots and Systems (IROS)}.\hskip 1em plus 0.5em
  minus 0.4em\relax IEEE, 2020, pp. 2007--2013.

\bibitem{bewley2016simple}
A.~Bewley, Z.~Ge, L.~Ott, F.~Ramos, and B.~Upcroft, ``Simple online and
  realtime tracking,'' in \emph{2016 IEEE international conference on image
  processing (ICIP)}.\hskip 1em plus 0.5em minus 0.4em\relax IEEE, 2016, pp.
  3464--3468.

\bibitem{liang2022neural}
M.~Liang and F.~Meyer, ``Neural enhanced belief propagation for data
  association in multiobject tracking,'' in \emph{2022 25th International
  Conference on Information Fusion (FUSION)}.\hskip 1em plus 0.5em minus
  0.4em\relax IEEE, 2022, pp. 1--7.

\bibitem{bai2022transfusion}
X.~Bai, \emph{et~al.}, ``Transfusion: Robust lidar-camera fusion for 3d object
  detection with transformers,'' in \emph{Proceedings of the IEEE/CVF
  Conference on Computer Vision and Pattern Recognition}, 2022, pp. 1090--1099.

\bibitem{carla}
A.~Dosovitskiy, G.~Ros, F.~Codevilla, A.~Lopez, and V.~Koltun, ``{CARLA}: {An}
  open urban driving simulator,'' in \emph{Proceedings of the 1st Annual
  Conference on Robot Learning}, 2017, pp. 1--16.

\bibitem{nuscenes2019}
H.~Caesar, \emph{et~al.}, ``nuscenes: A multimodal dataset for autonomous
  driving,'' \emph{arXiv preprint arXiv:1903.11027}, 2019.

\bibitem{thrun2002probabilistic}
S.~Thrun, ``Probabilistic robotics,'' \emph{Communications of the ACM},
  vol.~45, no.~3, pp. 52--57, 2002.

\bibitem{tanzmeister2014grid}
G.~Tanzmeister, J.~Thomas, D.~Wollherr, and M.~Buss, ``Grid-based mapping and
  tracking in dynamic environments using a uniform evidential environment
  representation,'' in \emph{2014 IEEE International Conference on Robotics and
  Automation (ICRA)}.\hskip 1em plus 0.5em minus 0.4em\relax IEEE, 2014, pp.
  6090--6095.

\bibitem{9561375}
M.~Schreiber, V.~Belagiannis, C.~Gl{\"a}ser, and K.~Dietmayer, ``Dynamic
  occupancy grid mapping with recurrent neural networks,'' in \emph{2021 IEEE
  International Conference on Robotics and Automation (ICRA)}, 2021, pp.
  6717--6724.

\bibitem{fernandez2019real}
M.~Fern{\'a}ndez-Sanjurjo, M.~Mucientes, and V.~M. Brea, ``A real-time
  processing stand-alone multiple object visual tracking system,'' in
  \emph{Computer Analysis of Images and Patterns: 18th International
  Conference, CAIP 2019, Salerno, Italy, September 3--5, 2019, Proceedings,
  Part I}.\hskip 1em plus 0.5em minus 0.4em\relax Springer, 2019, pp. 64--74.

\bibitem{kalman_filter}
R.~E. Kalman, ``A new approach to linear filtering and prediction problems,''
  1960.

\bibitem{hungarian_algorithm}
H.~W. Kuhn, ``The hungarian method for the assignment problem,'' \emph{Naval
  research logistics quarterly}, vol.~2, no. 1-2, pp. 83--97, 1955.

\bibitem{wojke2017simple}
N.~Wojke, A.~Bewley, and D.~Paulus, ``Simple online and realtime tracking with
  a deep association metric,'' in \emph{2017 IEEE international conference on
  image processing (ICIP)}.\hskip 1em plus 0.5em minus 0.4em\relax IEEE, 2017,
  pp. 3645--3649.

\bibitem{kim2021eagermot}
A.~Kim, A.~O{\v{s}}ep, and L.~Leal-Taix{\'e}, ``Eagermot: 3d multi-object
  tracking via sensor fusion,'' in \emph{2021 IEEE International Conference on
  Robotics and Automation (ICRA)}.\hskip 1em plus 0.5em minus 0.4em\relax IEEE,
  2021, pp. 11\,315--11\,321.

\bibitem{Meinhardt_2022_CVPR}
T.~Meinhardt, A.~Kirillov, L.~Leal-Taix\'e, and C.~Feichtenhofer,
  ``Trackformer: Multi-object tracking with transformers,'' in
  \emph{Proceedings of the IEEE/CVF Conference on Computer Vision and Pattern
  Recognition (CVPR)}, June 2022, pp. 8844--8854.

\bibitem{cid_pose}
D.~Wang and S.~Zhang, ``Contextual instance decoupling for robust multi-person
  pose estimation,'' in \emph{2022 IEEE/CVF Conference on Computer Vision and
  Pattern Recognition (CVPR)}, 2022, pp. 11\,050--11\,058.

\bibitem{coco_dataset}
T.-Y. Lin, \emph{et~al.}, ``Microsoft coco: Common objects in context,'' in
  \emph{Computer Vision--ECCV 2014: 13th European Conference, Zurich,
  Switzerland, September 6-12, 2014, Proceedings, Part V 13}.\hskip 1em plus
  0.5em minus 0.4em\relax Springer, 2014, pp. 740--755.

\bibitem{giou}
H.~Rezatofighi, N.~Tsoi, J.~Gwak, A.~Sadeghian, I.~Reid, and S.~Savarese,
  ``Generalized intersection over union,'' June 2019.

\bibitem{hrnet}
K.~Sun, B.~Xiao, D.~Liu, and J.~Wang, ``Deep high-resolution representation
  learning for human pose estimation,'' in \emph{Proceedings of the IEEE/CVF
  conference on computer vision and pattern recognition}, 2019, pp. 5693--5703.

\bibitem{pytorch}
A.~Paszke, \emph{et~al.}, ``Pytorch: An imperative style, high-performance deep
  learning library,'' in \emph{Advances in Neural Information Processing
  Systems 32}.\hskip 1em plus 0.5em minus 0.4em\relax Curran Associates, Inc.,
  2019, pp. 8024--8035.

\end{thebibliography}

\end{document}